\documentclass{article}

\usepackage{arxiv}

\usepackage[utf8]{inputenc} 
\usepackage[T1]{fontenc}    
\usepackage{hyperref}       
\usepackage{url}            
\usepackage{booktabs}       
\usepackage{amsfonts}       
\usepackage{nicefrac}       
\usepackage{microtype}      
\usepackage{comment}

\usepackage{amsmath}
\usepackage{blkarray, bigstrut}
\usepackage{float}
\usepackage{tikz}

\usetikzlibrary{calc,angles,patterns,patterns.meta,decorations.pathmorphing}

\title{Physical Constraint Embedded Neural Networks for inference and noise regulation}

\author{
  Gregory Barber, ~Mulugeta A.~Haile\thanks{Corresponding author: \href{mailto:mulugeta.a.haile.civ@mail.mil}{mulugeta.a.haile.civ@mail.mil}}, ~and ~Tzikang Chen \\
  U.S. Army Research Laboratory\\
  Aberdeen Proving Ground, MD 21005\\

}

\begin{document}
\maketitle

\begin{abstract}
\noindent

Neural networks often require large amounts of data to generalize and can be ill-suited for modeling small and noisy experimental datasets. Standard network architectures trained on scarce and noisy data will return predictions that violate the underlying physics. In this paper, we present methods for embedding even--odd symmetries and conservation laws in neural networks and propose novel extensions and use cases for physical constraint embedded neural networks. We design an even--odd decomposition architecture for disentangling a neural network parameterized function into its even and odd components and demonstrate that it can accurately infer symmetries without prior knowledge. We highlight the noise resilient properties of physical constraint embedded neural networks and demonstrate their utility as physics-informed noise regulators. Here we employed a conservation of energy constraint embedded network as a physics-informed noise regulator for a symbolic regression task. We showed that our approach returns a symbolic representation of the neural network parameterized function that aligns well with the underlying physics while outperforming a baseline symbolic regression approach. 

\end{abstract}

\section{Introduction}

Neural network performance is highly dependent on the quantity and quality of the available training data. Experimental datasets are unavoidably noisy and scarce in quantity. This can limit a network's ability to learn features and generalize. For neural networks modeling physical behaviors, this can be particularly problematic as it can lead to physically inconsistent predictions that violate governing laws. The black-box nature of the learned features and relations makes it challenging to assess if the network is accurately learning the underlying physics, constraints, and parameters from data. These issues of network's interpretability and generalization ability limit their utility to model and simulate physical systems.

The issues regarding interpretability of learned features and physical consistency of predictions have been active areas of research for the past few years in what is now broadly referred to as scientific machine learning \cite{baker2019workshop, chuang2018adversarial, rackauckas2020universal} and physics-aware or physics-informed neural networks \cite{zamzam2020physics, seo2019physics, raissi2019pinns, tartakovsky2018learning, mao2020physics, mcclenny2021tensordiffeq}. Some of these efforts have resulted in neural alternatives for modeling and simulation of simple dynamical systems \cite{saemundsson2020variational, greydanus2019} and for solving ordinary and partial differential equations \cite{ chen2018neural,raissi2019pinns,rackauckas2020universal, doi:10.1177/1475921719881642}.

Here we will discuss a subset of neural networks and modeling architectures that have shown promising results and present novel extensions and use cases. Our emphasis is on physical constraint embedded neural networks suitable for modeling and simulation of dynamical systems as well as neural models that can learn properties of physical systems. By physical constraint embedded neural networks, we are referring to neural architectures that respect an underlying physics as a result of their design and loss constraints. Our end goal is to realize a model architecture that can not only make physically consistent predictions but also can be incorporated into existing modeling approaches to improve generalization ability, regulate noise, and infer physical parameters \cite{barber2021joint}.

Physical constraint embedded networks learn an underlying structure in the data. This allows them to return more robust and generalizable predictions than standard neural architectures. There are two common approaches used to embed physical constraints in neural networks. The first involves structuring a network's architecture to ensure the desired symmetry and the second involves placing constraints in the loss function to encourage the model to learn symmetry. We will present examples of both approaches and show that they can achieve similar results for embedding a generic even or odd symmetry.

This paper is organized as follows. Section \ref{sec:sec2} deals with neural architectures for embedding even--odd symmetries and presents a novel even--odd decomposition network for decomposing a function into even and odd components. Section \ref{sec:sec3} deals with neural architectures for embedding conservation of energy constraint. Section \ref{sec:sec4} discusses symbolic regression and proposes the use of physical constraint embedded networks as physics-informed noise regulators for symbolic regression. Section \ref{sec:sec5} summarizes our findings and offers concluding remarks. 
\section{Even--Odd Symmetry}
\label{sec:sec2}

Conservation laws such as conservation of energy, momentum, mass, and electric charge are foundational concepts for understanding the behavior of classical physical systems and are intimately related to symmetry. Symmetry in physics refers to physical or mathematical features of a physical system that remain invariant during shifts in time or space. Even--odd symmetry refers to symmetry across the y-axis. An even function, such as such as $x^2$ and $\cos(x)$, displays mirror symmetry whereas an odd function, such as such as $x^3$ and $\sin(x)$, displays rotational symmetry. Given prior knowledge of a target system's underlying even or odd symmetry, it may prove useful to provide this information directly to a neural network. Mattheakis et al.\cite{mattheakis2019} proposed an even--odd hub neuron approach as a means of embedding these symmetries in neural networks. These hub neurons replaced the last hidden layer in a standard multilayer perceptron (MLP) and decomposed the network's output to return the desired even or odd component. This decomposition proved resilient to noisy data and outperformed a standard feed-forward network.

\subsection{Even--odd networks}

As a motivating example in embedding a physical constraint in a network, we will replicate the results from \cite{mattheakis2019} for a simple even function and train an additional network with a symmetry loss constraint. Here we will start with a derivation of the even--odd hub neuron approach. We will then modify this approach to remove the need for prior knowledge of the underlying symmetry. Lastly, we will extend this approach to perform a network-driven even--odd decomposition and show that it can decompose toy neither even nor odd functions into their even and odd components.


\textbf{Even--odd hub neuron derivation}.
A function $f(x)$ displays even symmetry if it satisfies the condition $f(x) = f(-x)$ and odd symmetry if it satisfies the condition $f(x) = -f(-x)$. If neither of these conditions is satisfied the function is composed of a mixture of even and odd components. The function can then be decomposed into its even and odd components. An even--odd hub neuron applies this decomposition to the last hidden layer in a standard MLP splitting its output into even and odd components and returning the component to enforce the desired symmetry. 

The output $\hat{x}(t)$ of the last hidden layer in a standard MLP with $N$ neurons is given by the function:
\begin{equation}
\hat{x}(t) = \sum^N_{i = 1} w_i h_i(t) + b
\end{equation}
Where $w_i$ is the weights, $b$ is bias, $t$ is the input and $h_i(t)$ is the output of the activation function of the prior neuron.

The even odd decomposition of $\hat{x}(t)$ yields:
\begin{equation}
\begin{split}
\hat{x}(t) & = \sum^N_{i = 1} w_i h_i(t) + b \\
\hat{x}(t) & = \underbrace{\frac{\hat{x}(t) + \hat{x}(-t)}{2}}_{even} + \underbrace{\frac{\hat{x}(t) - \hat{x}(-t)}{2}}_{odd} \\
\hat{x}(t) & = \underbrace{\frac{1}{2}\sum^N_{i=1}w_i[h_i(t)+h_i(-t)] + 2b}_{even} \   + \ \underbrace{\frac{1}{2}\sum^N_{i=1}w_i[h_i(t)-h_i(-t)]}_{odd}
\end{split}
\end{equation}

The output of an even hub neuron is then given by: 
\begin{equation}
\hat{x}(t) = \frac{1}{2}\sum^N_{i=1}w_i[h_i(t)+h_i(-t)] + 2b 
\end{equation}

and an odd hub neuron by:
\begin{equation}
\hat{x}(t) = \frac{1}{2}\sum^N_{i=1}w_i[h_i(t)-h_i(-t)]   
\end{equation}

\textbf{An even--odd metric} can be used to assess the model's ability to learn the underlying even or odd symmetry. For an even symmetry the mean squared error of $MLP(x)$ and $MLP(-x)$ can be used and for an odd symmetry the mean squared of $MLP(x)$ and $-MLP(-x)$ can be used. 

\begin{figure}[H]
    \includegraphics{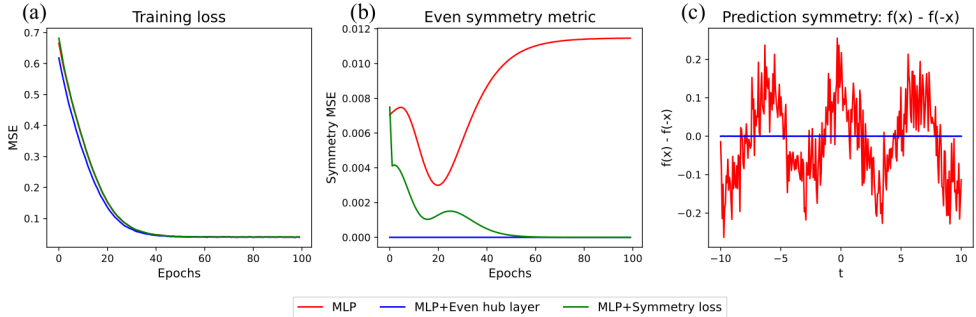}
    \centering
    \caption{Even symmetry results. (a) Training loss, all three models performed similarly during training. (b) Even symmetry metric during training. The MLP with the Even hub layer consistently returned symmetric results. The MLP with the symmetry loss term learned the symmetry during training and reduced the symmetry metric. The standard MLP failed to learn symmetry. (c) Prediction symmetry, after training positive and negative values for $x$ were passed through the networks and their differences were plotted. The two constraint informed networks, the MLP with the Even hub layer and the MLP trained with a Symmetry loss term, obeyed the symmetry and returned values close to zero. The standard MLP failed to do so and over-fit to the noise.}
    \label{fig:Even}
\end{figure}

\textbf{Implementation.}
A standard MLP and an MLP with an even--odd hub layer were implemented in PyTorch following the architecture described in \cite{mattheakis2019}. The standard MLP model consisted of two fully connected layers of 5 neurons each, with sigmoid activations and one fully connected output layer. The MLP with the even--odd hub layer replaced the last hidden layer with an even--odd hub neuron. As in \cite{mattheakis2019} we applied the models to a dataset generated from even (cosine) or odd (sine) functions with normally distributed noise $N \sim (\mu = 0,\sigma = 0.2)$. The models were trained to minimize the mean squared error (MSE) of $x(t)$ and $\hat{x}(t)$. This loss function was then modified to include the symmetry metric term and an additional standard MLP model was trained. Figure \ref{fig:Even} compares results for even symmetry function. 


\textbf{Results.} Three models were trained: a standard MLP, an MLP with an Even hub layer, and an MLP with symmetry loss term. The models all converged to similar loss values during training, Figure \ref{fig:Even}(a). The standard MLP initially reduced the even symmetry metric before diverging. The MLP with the Even hub layer constantly preserved the symmetry returning small symmetry metric values around 0 during training, Figure \ref{fig:Even}(b). The MLP with the symmetry loss term successfully learned the symmetry during training indicated by the decrease in its symmetry metric, Figure \ref{fig:Even}(b). Both of the constraints informed networks, the MLP with the Even hub layer and the MLP trained with symmetry loss term, returned predictions over the training interval that respected underlying symmetry. The standard MLP failed to do so and was prone to overfitting to the noise.   

\subsection{Even--odd decomposition network}

\begin{figure}[H]
    \includegraphics[height=2.2cm]{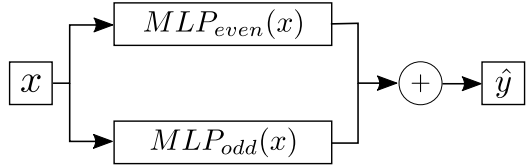}
    \centering
    \caption{Even--odd decomposition network. The model input $x$ is first passed through both an even network ($MLP_{even}$) and an odd network ($MLP_{odd}$). The output of both networks is then summed and taken as the prediction $\hat{y}$. A symmetry constraint, shown in Equation \ref{even_odd_loss}, is present in the loss function for both the even and odd networks.}
    \label{fig:decomp_arch}
\end{figure}

The even--odd hub neuron approach outperforms a standard feed-forward network and is resilient to noisy input data. This approach however suffers from limitations. It requires prior knowledge of a target system's even or odd symmetry and cannot be applied to systems that are a mixture of even and odd components. To address these limitations we designed an Even--odd decomposition network for cases where the underlying symmetry is assumed to be unknown or is a mixture of even and odd components. Figure \ref{fig:decomp_arch} shows the outline of the architecture.

\begin{figure}[H]
    \includegraphics[height=8cm]{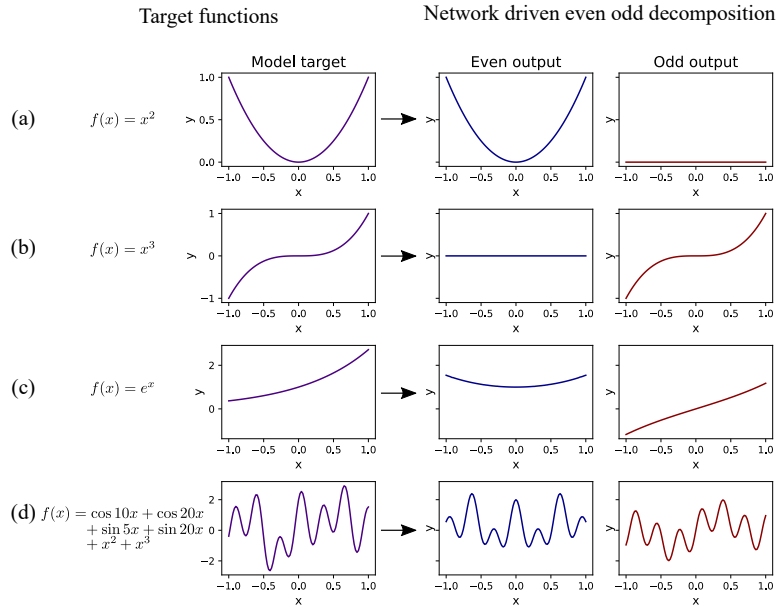}
    \centering
    \caption{Extending even--odd networks. In (a) an even function $x^2$ is targeted by the decomposition network where the network learns to only encode the even component, its odd component returns values close to 0. In (b) an odd function $x^3$ is targeted where the network only encodes the odd component while the even component returns values close to 0. In (c) a function that is neither even nor odd $e^x$ is targeted, here the network successfully decomposes the even and odd components. Its even component encodes the hyperbolic cosine and its odd component encodes the hyperbolic sine.  In (d) a function that is mixture of polynomial terms and trigonometric functions of varying frequencies is targeted, here the network successfully decomposes the even ($\cos{10x} + \cos{20x} + x^2$) and odd ($\sin{5x} + \cos{20x} + x^3 $) components.}
    \label{fig:decomp}
\end{figure}

The even--odd decomposition network shown in Figure \ref{fig:decomp_arch} consists of two components: an even component and an odd component. The even component contains an even network and seeks to parameterize the even portion of the underlying function. The odd component contains an odd network and seeks to parameterize the odd portion of the underlying function. As input, both the even and odd components receive the same $x$. The output for both networks is then summed to return the prediction $\hat{y}$. During training, the even--odd symmetries of the respective components are enforced through the loss function, Equation \ref{even_odd_loss}. Here in addition to prediction accuracy, we included a symmetry metric for the output of both the even and odd components. We evaluated the even--odd decomposition network on a selection of simple functions that display even and odd symmetries, as well as functions that are neither even nor odd, Figure \ref{fig:decomp}. 
\begin{equation}[H]
Loss = ||\hat{y} - y||_2 + ||\text{MLP}_{even}(x) -\text{MLP}_{even}(-x)||_2 +  ||\text{MLP}_{odd}(x) + \text{MLP}_{odd}(-x)||_2
\label{even_odd_loss}
\end{equation}
In Figure \ref{fig:decomp}, we present the network-driven even--odd decomposition results. The even--odd decomposition network was able to successfully learn the underlying symmetry without prior knowledge. Given an even target function, $x^2$ and odd target function $x^3$ the model learned to only encode the target function into the network's respective even or odd component, Figure \ref{fig:decomp}(a and b). For the even target function, the network's even component learned a parameterization of the target function and output its value while the network's odd component returned small values close to 0.  For the odd target function, the network's odd component learned a parameterization of the target function and output its value while the network's even component returned small values close to 0. An even or odd component returning values close to 0 indicates that it is not encoding information and suggests the target is solely an even or odd function. 

The even--odd decomposition network was able to successfully decompose a selection of neither even nor odd functions into their even and odd components. In Figure \ref{fig:decomp}(c) the target function $e^x$ was decomposed by the network into its even and odd components the hyperbolic cosine and sine. In Figure \ref{fig:decomp}(d) a target function containing polynomial terms and cosine and sine functions of varying frequencies was decomposed into its even and odd components. Decomposing a neural network parameterized function into its even and odd components may offer insight into the structure of the underlying function. This might have applications in equation discovery. For instance, an even--odd decomposition network could be used to split an unknown function into its even or odd components. These components could then be model separately with an equation discovery algorithm making use of the network-derived symmetry knowledge. The even--odd decomposition network is limited to centered data for non centered input data it does not guarantee the return of the simplest even or odd decomposition required to describe the data. Here a complexity metric may be needed to improve performance such as the number of terms present in a symbolic fit on the even or odd components.

In this section, we discussed methods for embedding even and odd symmetries in a neural network through both their architecture and loss function. We then demonstrated that these approaches outperformed a standard feed-forward network and extended them to allow inference into a network parameterized function's symmetry and even odd decomposition. In the next section, we will discuss embedding conservation of energy constraint.

\section{Conservation of Energy}
\label{sec:sec3}

The total energy of a dynamic system is given by the sum of its potential and kinetic energies. A system in which this value remains constant over time is said to conserve energy. The dynamics of an ideal energy-conserving system can be formulated with Hamiltonian mechanics. The Hamiltonian $\cal{H}$ is a function that relates a system's total energy to its canonical coordinates i.e. its generalized coordinates $\mathbf{q}$ and momentum $\mathbf{p}$.

The Hamiltonian of a system is given by:
\begin{equation}
E_{total} =\cal{H}(\mathbf{q},\mathbf{p})
\label{Ham}
\end{equation}
Where the total energy $E_{total}$ is the sum of the kinetic and potential energies. The time evolutions of the states are given by:
\begin{equation}
\begin{split}
\frac{d \mathbf{q}}{dt} & = \phantom{-}\frac{\partial \cal{H}}{\partial \mathbf{p}} \\ \frac{d\mathbf{p}}{dt} & = - \frac{\partial \cal{H}}{\partial\mathbf{q}}
\end{split}
\label{time_ev}
\end{equation}
From this formulation, the position and momentum can be evaluated at any time point through integration over time. Deriving the Hamiltonian formulation for a system from data can be challenging. Given data, it may be advantageous to parameterize a system's Hamiltonian and time evolutions with a physics embedded modeling framework. Neural networks act as universal function approximators \cite{hornik1991approximation} and as such, they are a natural choice for parameterizing a system's Hamiltonian and its time evolutions.

Standard feed-forward neural networks, however, can struggle generalizing to observations outside their training range, and neural network architectures widely used for time series modeling such as RNNs and LSTMs \cite{hochreiter1997long} suffer shortcomings for dynamical systems: they require $n$ previous time steps, their learned features and parameters are difficult to interpret, and they can drift for far-out predictions. Neural ordinary differential equations \cite{chen2018neural} and related approaches directly modeling time derivatives offer an attractive alternative for modeling continuous time series, however, they do not embed a physical constraint. Here we will discuss Hamiltonian neural networks a neural architecture that embeds conservation of energy constraint through structuring a network to parameterize a system's Hamiltonian and time evolutions.  

\textbf{Hamiltonian neural networks}. Hamiltonian mechanics was first utilized in neural network design nearly three decades ago~\cite{article_oldHNN} and in recent years this area has seen renewed interest \cite{greydanus2019, toth2020hamiltonian, Bertalan_2019,mattheakis2020hamiltonian, barber2021joint}. Hamiltonian neural networks (HNNs) are a neural architecture that learns conservation laws in an unsupervised manner \cite{greydanus2019}. HNNs operate by parameterizing the Hamiltonian function with a neural network.

\begin{figure}[H]
    \includegraphics[width=8cm]{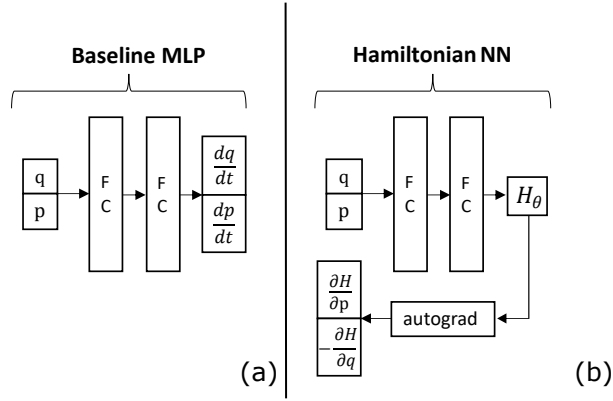}
    \centering
    \caption{HNN and Baseline MLP architecture. FC indicates a fully connected network layer, $H_\theta$ is Hamiltonian like value parameterized by the neural network. The Baseline MLP directly returns $\frac{dq}{dt},\frac{dp}{dt}$ while the HNN first returns a Hamiltonian like value and then computes it's partials to return the time evolutions.}
    \label{fig:HNN}
\end{figure}

A HNN requires observations of a system's canonical coordinates to train. This is generally provided to the network as input. If a system's canonical coordinates are unknown, high dimensional data such as images can be paired with dimensionality reduction approaches to extract a representation of a system's canonical coordinates. The Hamiltonian Generative Network architecture \cite{toth2020hamiltonian} has shown success in extracting representations of canonical coordinates and modeling classical physical systems from images. Here we will focus our discussion around the classic HNN architecture from \cite{greydanus2019} illustrated in Figure \ref{fig:HNN}(b), that receives only a system's canonical coordinates as input.

The canonical coordinates input to an HNN are first past through a feed forward neural network and an energy-like value $H_{\theta}$ is output from the network. This portion of the model is a parameterization of Equation \ref{Ham}:
\begin{equation}
H_{\theta} = \text{HNN}(\mathbf{q},\mathbf{p})
\label{hnn_eq}
\end{equation}
The partial derivatives of $H_{\theta}$ with respect to the input canonical coordinates are then computed with automatic differentiation. This portion of the model is a parameterization of Equation \ref{time_ev}:
\begin{equation}
\begin{split}
\frac{d \mathbf{q}}{dt} & = \phantom{-}\frac{\partial H_{\theta}}{\partial \mathbf{p}} \\ \frac{d\mathbf{p}}{dt} & = - \frac{\partial H_{\theta}}{\partial\mathbf{q}}
\label{hnn_pars}
\end{split}
\end{equation}
A HNN is trained to minimize the mean squared error metric between the partials derivatives of $H_{\theta}$ returned from the network Equation \ref{hnn_pars} and the time derivatives computed from the input canonical coordinates. This loss function is given in Equation \ref{hnn_loss}:
\begin{equation}
Loss = ||\frac{\partial H_\theta}{\partial p} - \frac{dq}{dt}||_2 + ||\frac{\partial H_\theta}{\partial q} + \frac{dp}{dt}||_2
\label{hnn_loss}
\end{equation}
After training a HNN can be evaluated with an ODE solver to obtain coordinate predictions. As a motivating example in embedding conservation of energy constraint through a HNN we will replicate the results \cite{greydanus2019} for a noisy ideal mass-spring system and discuss architectural choices.

\textbf{Implementation}. As in \cite{greydanus2019} we implemented two networks in PyTorch \cite{paszke2019pytorch} a Baseline MLP Figure \ref{fig:HNN}(a) and a HNN Figure \ref{fig:HNN}(b). The Baseline MLP directly outputs the time derivatives and served as a point of comparison for the constraint embedded HNN network. The Baseline MLP consisted of two fully connected layers of 200 neurons each with Tanh activation functions and one fully connected output layer of size 2 corresponding to $(\frac{dq}{dt}, \frac{dp}{dt})$. The HNN consisted of two equation parameterizations, Equation \ref{hnn_eq} and \ref{hnn_pars}. The former parameterized the Hamiltonian and consisted of two fully connected layers of 200 neurons each with Tanh activation functions and one fully connected output layer of size 1 corresponding to $H_{\theta}$. The latter parameterized the time evolutions and used the python Autograd library \cite{maclaurin2015autograd} to return the partials of $H_\theta$ with respect to $q$ and $p$, $(\frac{\partial H}{\partial p},-\frac{\partial H}{\partial q})$. After training an ODE solver was used to integrate both the Baseline and HNN models to generate coordinate predictions. The SciPy \cite{2020SciPy-NMeth} \texttt{solve\_ivp} function was used to integrate the network with a tolerance of 1e-12 and a 4th order Runge-Kutta method was used for the solver \cite{runge1895numerische}.

\begin{figure}[H]
    \includegraphics[width=10cm,height=4cm]{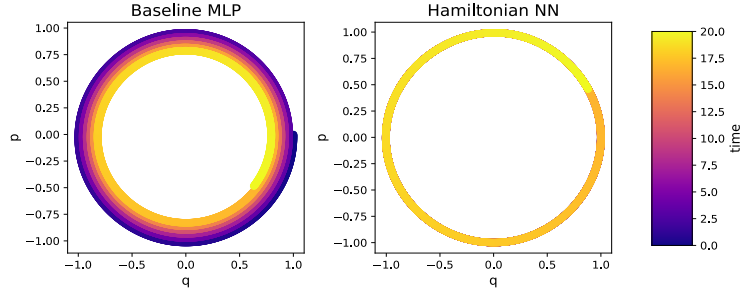}
    \centering
    \caption{Phase spaces predictions of a baseline MLP and a HNN; both trained on a noisy ideal mass-spring system. The HNN respects the conservation of energy constraint and conserves energy overtime. The Baseline MLP loses energy and spirals inward over time.}
    \label{fig:HNN_Phase}
\end{figure}

\textbf{Results}. We replicated the results of \cite{greydanus2019} for a noisy ideal mass-spring system. In Figure \ref{fig:HNN_Phase} we plot the coordinate predictions for each model over the time interval 0 to 20. As in \cite{greydanus2019} the HNN model learned to respect the conservation of energy constraint and returned physically consistent predictions. The HNN model was resilient to the noisy input and returned a Hamiltonian parameterization that generalized outside the training window. The baseline MLP model failed to generalize with respect to time and returned physically inconsistent predictions that lost energy over time. \cite{greydanus2019} generally reported results over a short time interval, t-span = [0,20] and reported results from HNNs using Tanh activation functions in their architecture. Expanding this evaluation interval to t-span = [0,500] and replacing the activation functions led to interplay between the structure of the network parameterization learned in the HNN and its ODE solver evaluation.

\begin{figure}[H]
    \includegraphics[width=10cm, height=4.1cm]{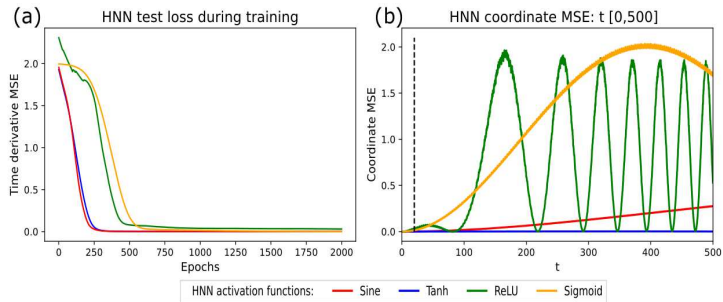}
    \centering
    \caption{HNN activation function choice. Four HNNs with different activation functions (ReLU, Tanh, Sigmoid, Sine) were trained on observations over the time interval [0,20] and evaluated over the time interval [0,500]. (a) The test loss (time derivative mean squared error) during training for the four models converged to a similar level over 2000 epochs. (b) Coordinate mean squared error of the ODE solver evaluations of the trained models. The black dotted line drawn at t = 20 demarcates the division between the training and extrapolation intervals. Despite achieving a similar test loss on the time derivatives during training the long-term coordinate predictions returned from the ODE solver evaluations varied greatly between the networks by activation function.}
    \label{fig:HNN_act}
\end{figure}

\textbf{HNN activation functions}. We found the choice of activation function used in a HNN impacts the model's long-term performance. In Figure  \ref{fig:HNN_act}(a), we present the test loss during training for four HNN models with varied activation functions and in Figure \ref{fig:HNN_act}(b), we present the ODE solver coordinate MSE for the four models evaluated after training over the interval t-span = [0,500]. The activation functions tested (ReLU, Tanh, Sigmoid) are widely used in deep learning and the (Sine) activation function has shown promising results in works with neural differential equations \cite{sitzmann2020implicit, mattheakis2020hamiltonian, saemundsson2020variational}.

All of the activation functions tested converged to a similar test loss during training. However, the ODE solver evaluations of these models were sensitive to the activation functions applied in the HNN. The Sigmoid and ReLU activation functions led to worse performance than the Sine or Tanh activation functions. A HNN's final test loss during training seems to be a poor indicator of its long-term predictive coordinate accuracy. Consideration in network design must then be made for the interplay between the Hamiltonian network parameterization learned in a HNN and the ODE solver evaluating it.

\textbf{Extending Hamiltonian neural networks.} The Hamiltonian neural network approach successfully embeds conservation of energy constraint in a neural network and returns network parameterization of a system's Hamiltonian. The Hamiltonian parameterization returned in a HNN is noise resilient to an extent and can extrapolate physically consistent predictions beyond the training interval. This noise resilient property of a HNN suggests it may have applications in noise regulation. In the next section, we will discuss symbolic regression and demonstrate a novel extension of an HNN as a physics informed tool for noise regulation, as well as utilize a symbolic regression approach to extract a symbolic representation of the network parameterized time evolution functions.

\section{Physical constraint embedded networks as a noise regulator for symbolic regression}
\label{sec:sec4}

Prior to this section, we have presented methods for embedding physical constraints into neural networks and neural network parameterizations of the underlying equations. In this section, we will present a data-driven method for discovering a system's underlying symbolic equations i.e. symbolic regression, and propose the use of physical constraint embedded neural works as noise regulators for symbolic regression.

Symbolic regression is a computationally demanding task, in which an algorithm attempts to build a candidate function from a provided library of possible operations, functions, and interactions e.g. ($-,+,\times,\div ,\sin x,e^{x}, x_1x_n$, etc). During this process, two factors are balanced: the number of terms in the candidate function and its predictive accuracy. A good symbolic fit will return a candidate function with the minimal numbers of terms required to achieve good predictive accuracy. As the complexity and number of input terms increase the computational search time required for a symbolic fit may become infeasible.
For generic functions, symbolic equation discovery can be nearly impossible \cite{udrescu2020ai}. However, for physical and dynamical systems commonalities and properties exist that can reduce the search space, increasing the chance of success. Symbolic regression approaches tend to capitalize on one or more of these properties.

Symbolic regression approaches included methods in genetic programming \cite{schmidt2009distilling}, sparse regression \cite{brunton2016discovering} and neural networks coupled with physics-inspired techniques \cite{udrescu2020ai}. Here we will focus our discussion around the sparse identification of nonlinear dynamics (SINDy) algorithm \cite{brunton2016discovering}, a sparse regression approach to symbolic regression. 

\subsection{Sparse identification of nonlinear dynamics}

The sparse identification of nonlinear dynamics (SINDy) algorithm introduced by \cite{brunton2016discovering} is a sparse regression approach to equation discovery in dynamical systems. This approach leverages the fact that most physical systems have only a few relevant terms that define the dynamics, making the governing equations sparse in a high-dimensional nonlinear function space \cite{brunton2016discovering}. 
In essence given observations of a system $\mathbf{X}$ and a library of all possible nonlinear functions $\mathbf{\Theta}$, the time evolution of the system $\mathbf{\dot{X}}$ depends on only a few active functions in the space $\mathbf{\Theta(X)}$. This setup can be framed as a sparse regression problem: $\mathbf{\dot{X}} = \mathbf{\Theta(X)}\mathbf{\Xi}$, where $\mathbf{\Xi}$ is a coefficient matrix determined in the regression whose nonzero terms indicate the active terms in $\mathbf{\Theta(X)}$. The sparse regression performed in the SINDy algorithm can utilize a selection of penalized regression optimizers including options such as LASSO \cite{tibshirani1996regression} and the sequential threshold least-squares (STLSQ) method proposed with SINDy \cite{brunton2016discovering}. These regression approaches bring with them a balance between equation complexity, accuracy and offer a computational time improvement over the brute force steps present in similar symbolic regression approaches. 

\begin{figure}[H]
\centering
\begin{tikzpicture}[scale=1.1]

\tikzstyle{mass} = [rectangle,fill=white,draw = black,inner sep=2mm]
\tikzstyle{spring} = [decoration={aspect=0.7, segment length= 5pt, amplitude= 4pt,coil},decorate, thick]

\node (a) [mass] at (0,0.6) {$m$};
\draw[decoration={aspect=0.7, segment length= 5pt, amplitude= 4pt,coil},decorate, thick] (0,2) -- (a);
\node[] at (.5,1.5) {$k$};

\fill [pattern = north east lines] (-1.5,2) rectangle (1.5,2.25);
\draw[thick] (-1.5,2) -- (1.5,2);
\coordinate (a1) at (-0.6,-.1);
\coordinate (a2) at (-0.6,1.3);
\draw[<->, thick] (a1) -- (a2);
\node[] at (-.9,0.6) {$q$};
\node[] at (0,2.5) { Ideal mass-spring};
\end{tikzpicture}
\caption{Ideal mass-spring where $m$ is the mass, $k$ is the spring coefficient and $q$ is the displacement from the neutral position.}
\label{sys_diagram}
\end{figure}
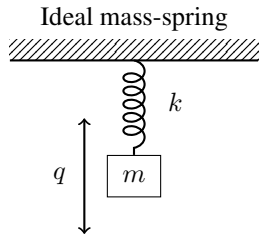
\vskip -.2in
As a motivating example, let us consider the dynamics of the mass-spring system shown in Figure \ref{sys_diagram} and also discussed in Section \ref{sec:sec3}, with the mass $m$ and spring constant $k$ set to $1$. The Hamiltonian for this system is given in Equation \ref{spring_ham}, for visualization purposes in the results we have dropped a factor of $\frac{1}{2}$ from the Hamiltonian formulation.
\begin{equation}
\mathcal{H} = \frac{p^2}{m} + kq^2
\label{spring_ham}
\end{equation}
The time evolution of this system is given by:
\begin{equation}
\begin{split}
& \dot{q} = \phantom{-}2p \\ 
& \dot{p} = -2q  
\end{split}
\label{spring_m_dxdt}
\end{equation}

The SINDy algorithm discovers the time evolution from observations of the system's canonical coordinates ($q, p$) and a library of possible functions. For this simple noiseless example SINDy was provided 5,000 coordinate observation and a library of possible functions limited to polynomials up to the 2nd degree. The solution to the regression problem $\mathbf{\dot{X}} = \mathbf{\Theta(X)}\mathbf{\Xi}$ for this system is given by:  
\begin{equation}
\rule{0mm}{11mm}\raisebox{2mm}{$\underbrace{\begin{bmatrix} \dot{q}(t_1) & \dot{p}(t_{1})  \\ 
                                                \dot{q}(t_2) & \dot{p}(t_{2}) \\ 
                                                \dot{q}(t_3) & \dot{p}(t_{3} \\ 
                                                \vdots & \vdots \\ 
                                                \dot{q}(t_n) & \dot{p}(t_{n})

\end{bmatrix}}_\mathbf{\textstyle \dot{X}}$}  = 
\raisebox{2mm}{$\underbrace{\begin{bmatrix} q(t_1) & p(t_1) & q^2(t_1) & p^2(t_1) \\ 
                                q(t_2) & p(t_2) & q^2(t_2) & p^2(t_2) \\
                                q(t_3) & p(t_3) & q^2(t_3) & p^2(t_3) \\
                                  \vdots & \vdots & \vdots & \vdots\\
                                q(t_n) & p(t_n) & q^2(t_n) & p^2(t_n)
                            
\end{bmatrix}}_\mathbf{\textstyle \Theta(X)}$}
\raisebox{5mm}{$\underbrace{\begin{bmatrix} 0 & -2  \\ 
                                            2 & \phantom{-}0 \\ 
                                            0 & \phantom{-}0 \\
                                            0 & \phantom{-}0

\end{bmatrix}}_{\textstyle \mathbf{\Xi}}$}
\end{equation}
Here the sparse regression returned a $\mathbf{\Xi}$ with two nonzero coefficients, corresponding to the coefficients for the terms $q$ and $p$ in $\mathbf{\Theta(X)}$. A symbolic equation for each column $k$ in $\mathbf{\dot{X}}$ can then be constructed using $\mathbf{\dot{x}_k} = \mathbf{\Theta(x^T)}\xi_k$: 
\begin{equation}
\begin{split}
& \dot{q} = \dot{x}_1 = \mathbf{\Theta(x^T)}\xi_1 = \phantom{-}2p \\
& \dot{p} = \dot{x}_2 = \mathbf{\Theta(x^T)}\xi_2 = -2q
\end{split}
\end{equation}
\textbf{SINDy with noise.} SINDy receives a systems coordinates $\mathbf{X}$ and their derivatives $\mathbf{\dot{X}}$ as input. If $\mathbf{\dot{X}}$ is unknown it can be approximated from $\mathbf{X}$ with methods such as the finite difference. This works well for noiseless cases as in the mass-spring example. However, introducing noise into this system can hurt SINDy's performance. This is illustrated in Figure \ref{fig:SINDy_phase_Noies}, with the introduction of noise the baseline SINDy models fails to extrapolate past zero. Noise reduction approaches applied to $\mathbf{X}$  such as smoothing or the total variation regularization \cite{chartrand2011numerical} numerical differentiation method applied in \cite{brunton2016discovering} can improve SINDy's performance on noisy data. These approaches however do not utilize prior knowledge of a systems physical constraint in noise regulation. In the next section we will discuss the use of physical constraint embedded neural networks as an additional tool for regulating noisy input for SINDy.

\noindent
\textbf{Deep learning with SINDy.} The SINDy algorithm has shown promising results when paired with deep learning neural architectures. It has been used to simultaneously learn the governing equations and the associated coordinate system \cite{champion2019data} and has been paired with neural network parameterized differential equations to discover unknown parts or the entirety of a governing equation \cite{rackauckas2020universal}. 

\subsection{SINDy+HNN approach for noise regulation}

\begin{figure}[H]
    \includegraphics[height=4cm]{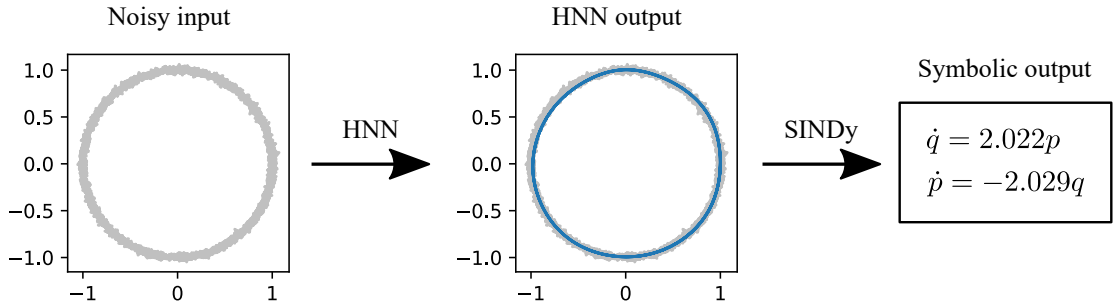}
    \centering
    \caption{Hamiltonian neural networks as a noise regulator for SINDy. A HNN is first trained on noisy coordinate data. The HNN model is then evaluated to return constraint informed coordinate predictions. Lastly, a SINDy model is fit to the HNN coordinates predictions and a symbolic representation of the HNN parameterized time evolution functions is returned.}
    \label{fig:SINDy_phase_flow}
\end{figure}

Symbolic regression approaches, for the most part, fit a model to data without regard to prior knowledge of a system's underlying physical constraints. As in unconstrained neural networks, this can cause issues for extrapolation. In the prior sections, we demonstrated that physical constraint embedded neural architectures were able to avoid this issue to an extent and were successful in filtering out noise. Here we will demonstrate that this property of physical constraint embedded networks can also be used as a physics-informed noise regulator to improve symbolic regression, in particular, we focus on using a HNN to improve the SINDy algorithm's results for the noisy mass-spring system given in Equation \ref{spring_m_dxdt}. We call this combined approach SINDy+HNN. 

Our aim here is not to suggest physical constraint embedded networks as a direct replacement for standard noise reduction methods such as smoothing, rather we aim to highlight their utility as an additional physics informed tool for noise regulation. As both the SINDy and HNN modeling approaches can benefit from standard noise reduction approaches and as our focus is on physical constraint embed networks for noise regulation we will forgo the standard noise reduction approaches and instead place the full burden of noise reduction on the HNN. 

\textbf{Methods.} 5,000 noisy observations for this system were generated from a starting position of $y_0 = [1,0]$. The noise in this system was sampled from a normal distribution centered around a mean $\mu = 0$. The noise level in this system was increased between each trial with the standard deviation of the noise distribution taking the following values: $\sigma = [0,0.01,0.02,0.03]$. Given this experimental setup, we will attempt to rediscover equations for the systems time evolutions $\dot{q}$ and $\dot{p}$. 

The noisy observation data was used to train two models for each noise level: a HNN using the architecture from Section \ref{sec:sec3} and a baseline SINDy model. After training, each HNN model was used to generate 5000 coordinate predictions. These predictions were then used to fit a SINDy+HNN model for each noise level. In total three models were returned for each noise level: a SINDy Baseline, a HNN, and a SINDy+HNN model. The SINDy+HNN approach is illustrated in Figure \ref{fig:SINDy_phase_flow}. For this example, the proposed SINDy+HNN approach to noise regulation is data-efficient requiring only a small data set of 5k observations.

All SINDy models in this work were implemented using the PySINDy python library \cite{desilva2020}. For each SINDy implementation, the same feature libraries were provided: a polynomial library containing terms up to the 2nd degree and a Fourier library limited to 1 frequency and containing both sine and cosine terms. The only argument that varied between the SINDy models was the threshold parameter provided to the STLSQ optimizer. Varying this parameter was necessary to ensure model performance and prevent the complete dropout of the terms in one model. For each model, the threshold parameter was selected as the value that dropped out the largest number of terms while maintaining the quality of the fit. Each fit was then integrated with an ODE solver to return the coordinate predictions. The same solver setup as in section 3 was used.

\begin{figure}
    \includegraphics[width=11cm,height=7cm]{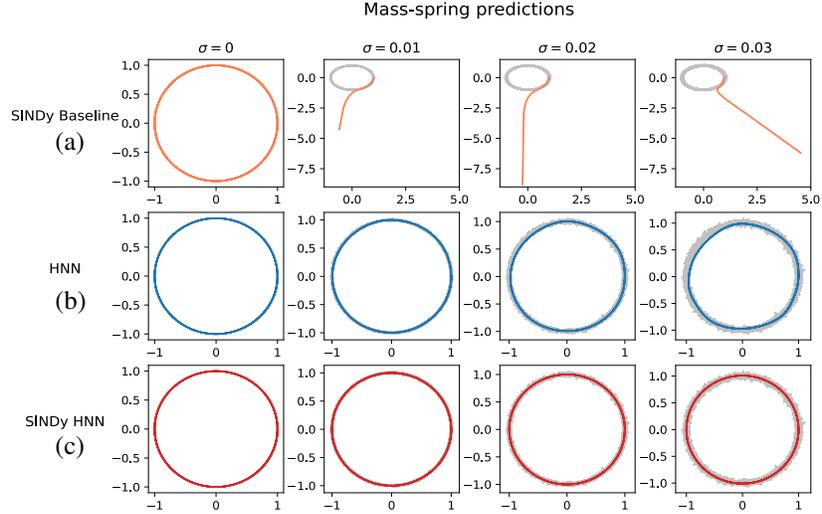}
    \put(-293,140){(a)}
    \put(-293,80){(b)}
    \put(-293,25){(c)}
    \centering
    \caption{Mass-spring phase trajectory predictions over the interval: t-span = [0,10]. The noise level increases from left to right: $\sigma = 0$ to $\sigma  = 0.03$. The noisy input data is plotted as a grey background trajectory. (a) The Baseline SINDy model fit directly on noisy input data. The baseline SINDy model struggles with the introduction of noise. (b) HNN model trained on noisy data. The HNN model learns to conserve energy however it begins to wobble at higher noise levels. (c) SINDy+HNN model, a SINDy model fit on HNN predictions. The SINDy+HNN model recovered equation close to the underlying equation as shown in Table \ref{table:Sym_results} and avoided the wobbling issue present at higher noise levels in the HNN model.}
    \label{fig:SINDy_phase_Noies}
\end{figure}

\textbf{Results.} The resulting phase space trajectory predictions for the three models are presented in Figure \ref{fig:SINDy_phase_Noies} and the corresponding symbolic results for each SINDy model are presented in Table \ref{table:Sym_results}. For all noise levels, the HNN and SINDy+HNN models outperformed the SINDy baseline model. This indicates that the conservation of energy constraint in the HNN is successful in reducing the impact of noise on the model. The HNN alone however is outperformed by the SINDy+HNN model. This is apparent in Figure  \ref{fig:SINDy_phase_Noies}(b) where the HNN's phase trajectory predictions wobble at higher noise levels.

The performance discrepancy between the HNN and SINDy+HNN models is a point of interest as the SINDy+HNN is extracting a symbolic representation of the neural network parameterized function in the HNN. The difference in their performance suggests that the neural network parameterized function may contain additional "weakly" contributing terms, perhaps as a result of a slight over-fit. These weakly contributing terms are then dropped in the thresholded SINDy fit. 

The SINDy+HNN approach successfully regulated the noisy input and recovered symbolic equations inline with the underlying equations Table \ref{table:Sym_results}. The SINDy Baseline model in contrast lacked a noise mitigation approach and was unable to recover meaningful symbolic equations when faced with noisy input data. As previously discussed both models could be improved by the inclusion of a standard noise reduction approach on the input coordinates. The strong performance of the SINDy+HNN in the absence of a standard noise regulation approach suggests an HNN may offer utility as a physics informed noise regulator. In practical applications, it may be advantageous to follow up a standard noise reduction approach with a physical constraint embedded network.

\begin{figure}[H]
    \includegraphics[width=11cm,height=7cm]{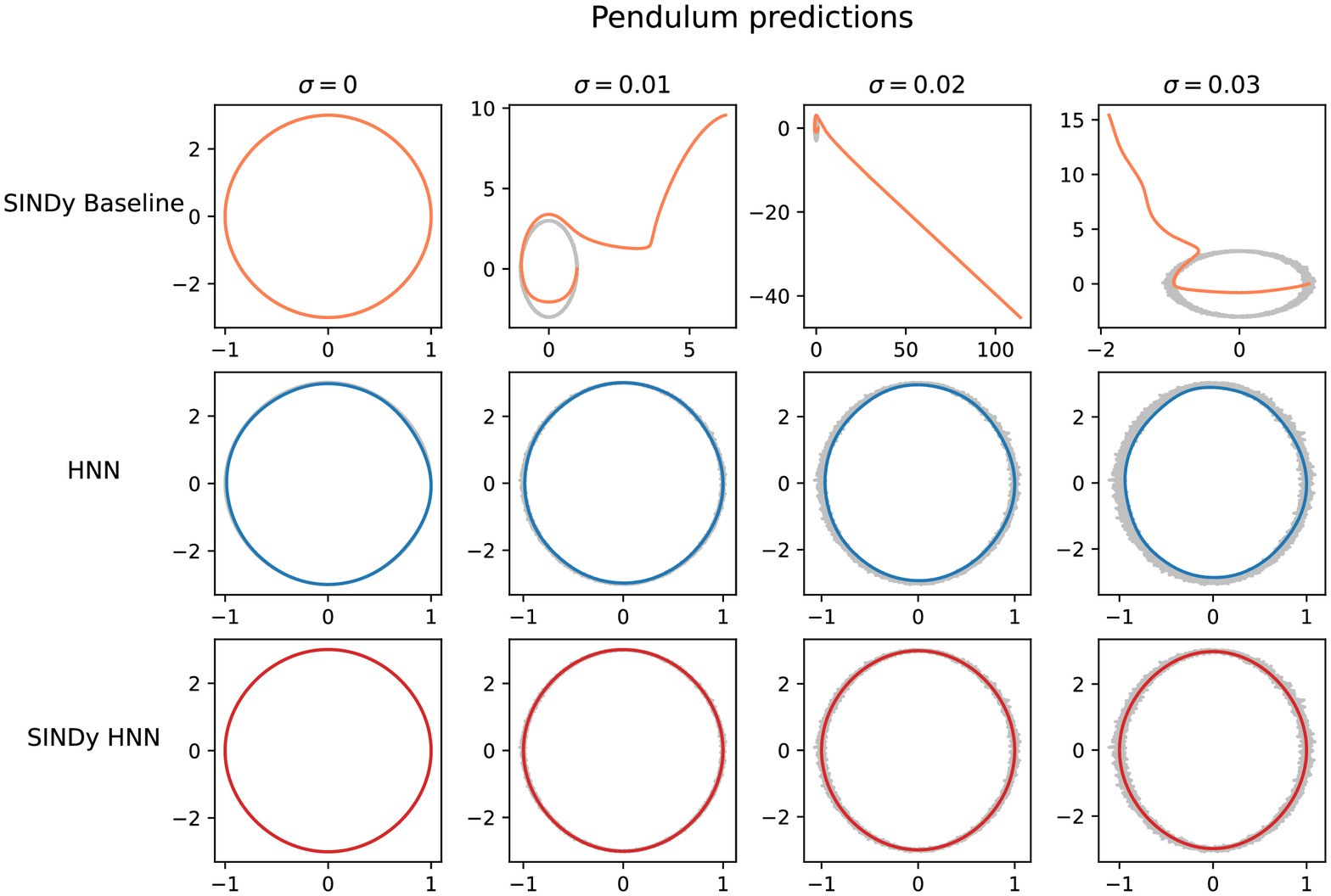}
    \put(-320,160){(a)}
    \put(-320,95){(b)}
    \put(-320,30){(c)}
    \centering
    \caption{Pendulum phase trajectory predictions over the interval: t-span = [0,10]. The noise level increases from left to right: $\sigma = 0$ to $\sigma  = 0.03$. The noisy input data is plotted as a grey background trajectory. (a) The Baseline SINDy model fit directly on noisy input data. The baseline SINDy model struggles with the introduction of noise. (b) HNN model trained on noisy data. The HNN model learns to conserve energy however it begins to wobble at higher noise levels. (c) SINDy+HNN model, a SINDy model fit on HNN predictions. The SINDy+HNN model recovered equation close to the underlying equation table: \ref{table:Sym_results_pen} and avoided the wobbling issue present at higher noise levels in the HNN model.}
    \label{fig:SINDy_pen_Noies}
    
\end{figure}

In Figure \ref{fig:SINDy_pen_Noies} and Table \ref{table:Sym_results_pen} we display the results of a baseline SINDy and SINDy+HNN approach for a noisy ideal pendulum. This trial followed the same procedure as outlined in the ideal mass-spring trial. The phase space trajectory predictions for the pendulum Figure \ref{fig:SINDy_pen_Noies} displays similar results to the mass-spring trial with the HNN and SINDy+HNN approaches outperforming the SINDy Baseline model on noisy data. The SINDy+HNN approach returned  symbolic equations in line with or close to the underlying equation Table \ref{table:Sym_results_pen}. As in the mass-spring trial, the SINDy baseline model was unable to recover meaningful symbolic equations when provided noisy input data.   

\begin{table}
\centering 
\caption{The Symbolic results returned by SINDy for a given noise level $\sigma$. Each row corresponds to a column in Figure \ref{fig:SINDy_phase_Noies} with the respective noies level.}
\renewcommand{\arraystretch}{1.3}
\begin{tabular}{clc}
\multicolumn{3}{c}{Symbolic results} \\
\toprule
$\sigma$&SINDy Baseline& SINDy+HNN \\[0.5ex] 
\midrule
$0$& $\begin{array} {lcl} \dot{q} & = & \phantom{-}1.999p \\ 
                          \dot{p} & = & -1.999q \end{array}$ & 
                          $\begin{array} {lcl} \dot{q} & = & \phantom{-}2.000p \\ 
                          \dot{p} & = & -1.999q \end{array}$\\ 
\hline
$0.01$& $\begin{array} {lcl} \dot{q} & = & -5.521q -14.016q^2 -7.564p^2 + \\ & & \phantom{-}6.068 \sin(q) + 14.040 \cos(p) \\ 
                          \dot{p} & = & -8.353p -4.764p^2 + 6.857\cos(q) + \\ & & \phantom{-}9.437\sin(p) -3.837 \cos(p) \end{array}$ & 
                          $\begin{array} {lcl} \dot{q} & = & \phantom{-}2.004p \\ 
                          \dot{p} & = & -2.003q \end{array}$\\ 
\hline
$0.02$& $\begin{array} {lcl} \dot{q} & = & -25.093q -20.579q^2 + 27.993\sin(q) + \\ & & -15.724 \cos(q) + 29.130\cos(p)\\ 
                          \dot{p} & = & \phantom{-}27.131 1 -26.288 p -18.102 p^2 + \\ & & \phantom{-} 7.986 \cos(q) + 29.449 \sin(p) 
                          \\ & & - 31.675 \cos(p) \end{array}$ & 
                          $\begin{array} {lcl} \dot{q} & = & \phantom{-}2.022p \\ 
                          \dot{p} & = & -2.029q \end{array}$\\ 
\hline
$0.03$& $\begin{array} {lcl} \dot{q} & = & -56.487q -19.686 q^2 + 63.174 \sin(q) + \\ & & -15.083 \cos(q) + 27.934 \cos(p) \\ 
                          \dot{p} & = & \phantom{-}25.206 -54.846p + 155.717q^2 \\ & & -172.541p^2 + 346.476\cos(q) + \\ & & 
                          \phantom{-} 61.323\sin(p) -368.441\cos(p) \end{array}$ & 
                          $\begin{array} {lcl} \dot{q} & = & \phantom{-}2.101p \\ 
                          \dot{p} & = & -2.139q \end{array}$\\ 
\bottomrule
\end{tabular}
\label{table:Sym_results}
\end{table}

\begin{table}[H]
\centering 
\caption{The Symbolic results returned by SINDy for a given noise level $\sigma$. Each row corresponds to a column in Figure \ref{fig:SINDy_pen_Noies} with the respective noies level.}
\renewcommand{\arraystretch}{1.3}
\begin{tabular}{cll}
\multicolumn{3}{c}{Pendulum symbolic results} \\
\toprule
$\sigma$&SINDy Baseline& SINDy+HNN \\[0.5ex] 
\midrule
$0$& $\begin{array} {lcl} \dot{q} & = & \phantom{-}1.000p  \\ 
                          \dot{p} & = & -9.795\sin(q)  \end{array}$ & 
                          $\begin{array} {lcl} \dot{q} & = & \phantom{-}0.994 p \\ 
                          \dot{p} & = & -9.804 \sin(q)   \end{array}$\\ 
\hline
$0.01$& $\begin{array} {lcl} \dot{q} & = &  \phantom{-}18.671 - 5.705q  - 1.892q^2 + 1.346p^2 +\\ & & \phantom{-}6.283\sin( q )  - 31.127\cos( q )   \\ 
                          \dot{p} & = & -9.798\sin(q)\end{array}$ & 
                          $\begin{array} {lcl} \dot{q} & = & \phantom{-}1.000 p  \\ 
                          \dot{p} & = & -9.869 \sin(q)   \end{array}$\\ 
\hline
$0.02$& $\begin{array} {lcl} \dot{q} & = & -12.606 - 26.049 q + 1.005 p + 12.680 q^2 + \\ & & \phantom{-}1.417 p^2 + 29.202\sin(q)\\ 
                          \dot{p} & = & \phantom{-}10.774+ 1.313 q  - 4.991 q^2  - 11.286 sin( q )  + \\ & & - 10.803 cos( q )   
                          \end{array}$ & 
                          $\begin{array} {lcl} \dot{q} & = & \phantom{-}1.045 p\\ 
                          \dot{p} & = &  -10.198\sin(q)  \end{array}$\\ 
\hline
$0.03$& $\begin{array} {lcl} \dot{q} & = & -389.390 - 56.961 q + 1.005 p + 186.087 q^2 + \\ & & - 0.924 p^2 + 63.872\sin(q) + 390.068 \cos(q) + \\ & & - 7.839\cos(p)  \\ 
                          \dot{p} & = & \phantom{-}22.086+ 2.452q  - 10.225q^2 - 12.574\sin(q)  + \\ & & - 22.156\cos(q)\end{array}$ & 
                          $\begin{array} {lcl} \dot{q} & = & \phantom{-} 1.121p   \\ 
                          \dot{p} & = & \phantom{-}6.356 q - 17.790\sin(q) \end{array}$\\ 
\bottomrule
\end{tabular}
\label{table:Sym_results_pen}
\end{table}

\section{Conclusion}
\label{sec:sec5}
We presented methods for embedding even--odd symmetries and conservation of energy in neural networks and introduced novel extensions and use cases for physical constraint embedded networks. We designed an even--odd decomposition network approach for disentangling a neural network parameterized function into its even and odd components. We evaluated this approach on centered data generated from a selection of even, odd, and neither even nor odd functions. Here we demonstrated that our approach successfully inferred the even or odd symmetry without prior knowledge of the underlying symmetry and decomposed neither even nor odd functions into their even and odd components. We also demonstrated the noise resilient properties of physical constraint embedded neural networks and showed that they could be utilized as physics-informed noise regulators for symbolic regression. Here we utilized a conservation of energy embedded network as a preprocessing noise filter for a SINDy approach. We evaluated this approach on two simple noisy dynamic systems. Our physical constraint informed noise regulation approach returned symbolic equations for the systems time evolutions in line with the underlying equations and outperformed a baseline model. We believe that extending and incorporating physical constraint embedded neural networks into model architectures can provide a strong tool for inference and physics-informed noise reduction.

\bibliographystyle{unsrt}

\bibliography{refs}

\end{document}